\documentclass{Interspeech2024}

\interspeechcameraready

\usepackage{float}
\usepackage{url}

\title{Unveiling Biases while Embracing Sustainability: Assessing the Dual Challenges of Automatic Speech Recognition Systems}

\name[affiliation={1,2}]{Ajinkya}{Kulkarni}
\name[affiliation={3}]{Atharva}{Kulkarni}
\name[affiliation={4,5}]{Miguel}{Couceiro}
\name[affiliation={5}]{Isabel}{Trancoso}

\address{
  $^1$IDIAP, Switzerland, 
  $^2$MBZUAI, UAE, $^3$Erisha Labs, India\\
  $^4$Université de Lorraine, CNRS, LORIA, Nancy, France \\
  $^5$INESC-ID, IST, Universidade de Lisboa,  Portugal}
\email{ajinkya.kulkarni@idiap.ch}


\begin{document}

\maketitle

\begin{abstract}

In this paper, we present a bias and sustainability focused investigation of Automatic Speech Recognition (ASR) systems, namely Whisper and Massively Multilingual Speech (MMS), which have achieved state-of-the-art (SOTA) performances. Despite their improved performance in controlled settings,
there remains a critical gap in understanding their efficacy and equity in real-world scenarios. We analyze ASR biases w.r.t. gender, accent, and age group, as well as their effect on downstream tasks. In addition, we examine the environmental impact of ASR systems, scrutinizing the use of large acoustic models on carbon emission and energy consumption. We also provide insights into our empirical analyses, offering a valuable contribution to the claims surrounding bias and sustainability in ASR systems.
    
    
\end{abstract}

\section{Introduction}

The advent of large deep neural networks (DNNs) has brought about substantial advancements in various speech-processing applications, notably in speech recognition. However, amidst this progress, there remains a notable gap in understanding the inherent biases towards gender, age groups, and accents. For instance, home assistant devices often exhibit biased performances towards non-native English speakers \cite{rac4}, limiting access to technology for certain individuals, particularly in speech-enabled human-machine interfaces \cite{bias_voice_assist1, bias_voice_assist2}. This exclusionary behavior may impede the usability of crucial services, such as emergency assistance for the elderly or navigation aids for differently-abled individuals. Thus, comprehensive studies of these large DNN models are crucial to ensuring their widespread and inclusive use.

In recent years, there has been a growing research community dedicated to examining biases in automatic speech recognition (ASR) systems, more specifically for English \cite{asr_biases_eng1,asr_biases_eng2,asr_biases_eng3}. This research primarily focuses on evaluating the disparities related to gender, age, accent, dialect, and racial attributes \cite{LimaFFA19, BlodgettBDW20, rac1,fr_bias1,fr_bias2,ar_bias1, mex_es_bias}. However, the training of large DNNs with extensive datasets necessitates ever-increasing computational resources, directly contributing to carbon emissions \cite{Strubell2019EnergyAP}. This environmental impact extends even to the inference time of these large DNN systems. The substantial release of CO2 into the atmosphere poses a significant threat to life on Earth, with consequences often overlooked within the deep learning research community, where effective benchmarking is lacking. These emissions not only disrupt the delicate balance of ecosystems but also raise profound ethical concerns regarding our responsibility towards the environment. Hence, investigating the carbon footprint and energy consumption of deep learning models is imperative for the sustainable development of deep learning systems.  In 2018, the Intergovernmental Panel on Climate Change emphasized the importance of limiting global temperature rise to below 1.5°C to mitigate adverse impacts on various aspects such as extreme weather events, ecosystems, and carbon removal efforts \footnote{\url{https://www.ipcc.ch/sr15/download/}}. To address this concern, efforts have been made to estimate carbon emissions and energy consumption. The Experimental Impact Tracker library, published in 2019 \cite{exp_impact_tracker}, was one such initiative. Subsequently, other carbon footprint tracking platforms like codecarbon\footnote{\url{https://codecarbon.io/}}, carbontracker\footnote{\url{https://carbontracker.org/}}, and eco2ai \cite{eco2ai} have emerged, providing support for tracking activity across GPU, CPU, user interfaces, and seamless integration into Python scripts. These tools enable organizations to compute carbon emissions and energy consumption, facilitating the development of sustainable AI systems.

The majority of studies on quantifying bias in ASR systems revolve around training individual systems and analyzing the disparities in ASR performance across different bias categories. For example, studies such as \cite{du_bias1,du_bias2} conducted bias analysis for Dutch using the Hidden Markov Model-DNN ASR system to observe gender bias. Subsequently, techniques such as vocal tract length normalization and data augmentation were proposed to mitigate biases in gender and age groups \cite{du_bias3}. Literature also suggests that ASR systems can exhibit bias at various stages of development, including data curation, model architecture design, and evaluation protocols \cite{Du2019FairnessID}. In 2018, an article titled "The Accent Gap" published in the Washington Post \cite{rac4} illustrated inconsistent performance across accents on commercial home assistant systems for non-native English speakers. Similarly, gender and accent analyses on YouTube's captions were described in \cite{rac2,rac3}, emphasizing the need for sociolinguistically-stratified validation of ASR systems. More recently, \cite{port1} established biases in multilingual ASR systems for Portuguese. Despite these efforts, there remains a lack of comprehensive studies focused on larger ASR systems for English.

\begin{figure*}[!t] 
  \centering
  \includegraphics[scale=0.074]{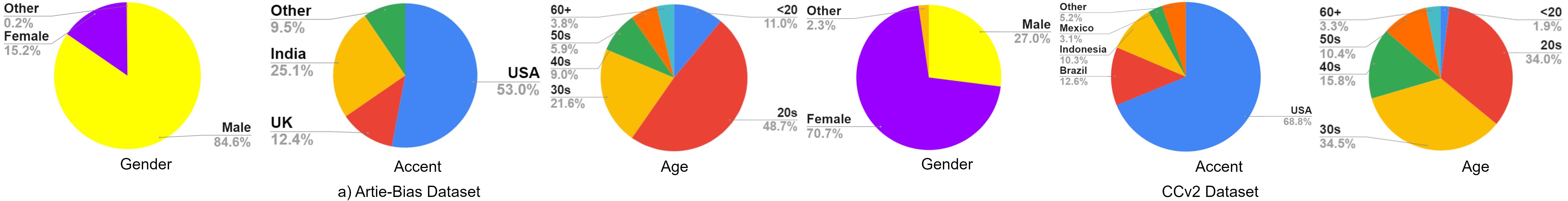} 
  \caption{Speech utterance distribution across gender, accent, and age for Artie-Bias and CCv2 dataset.}
  \label{fig:ccdv2_counts}
  \vspace{-1.5em}
\end{figure*}

ASR systems have progressively advanced to support over 100 languages \cite{asr2k,whisper,mms,ggleusm}. However, with the growing size of models and the vast amount of training data, computational requirements are also escalating, leading to increased carbon footprint and energy consumption. Research presented in \cite{Parcollet2021TheEA} detailed the training cost of ASR systems in terms of energy and carbon footprint, alongside improved performance. However, it is also essential to measure the carbon footprint during inference, given the widespread use of ASRs in both industry and society. Once deployed in real-world scenarios, it becomes challenging to backtrack, underscoring the importance of making users aware of their environmental impact and guiding them in selecting systems that align with their needs \cite{carbon_emit1,carbon_emit2}.

This paper addresses the dual challenges within English language ASR systems, focusing on potential biases and carbon footprint. We examine two recently proposed ASR variants, namely, the MMS \cite{mms} and Whisper \cite{whisper} ASR models. Our empirical study delves into two fairness-centric evaluation datasets, the Artie-bias dataset \cite{artiebias} and the Casual conversation dataset version 2 \cite{ccdv2}, which feature variations of read and spontaneous speech, respectively. To evaluate their energy consumption and carbon footprint, we employ three different carbon tracking libraries, across four distinct NVIDIA GPU systems. This approach allows for an integrated analysis of the impacts of using various GPU systems on different ASR models. The key contributions of this paper are as follows: \textbf{1.} We assess different variants of Whisper and MMS ASR systems on two bias-focused datasets, unveiling hidden disparities between read and spontaneous speech regarding gender, accent, and age. \textbf{2.} We present the first systematic sustainability study that not only compares different large ASR systems but also benchmark carbon tracking tools with various GPU variants.


	

Our findings reveal that for English, and easily replicable in other languages, {\it the Whisper variants perform better than MMS} on read speech for all three categories (accent, age, and gender). However, there is a {\it drastic performance degradation of Whisper in spontaneous speech}. 
For individual categories, we observed significant performance differences between the two types of models. 
Also, it was surprising to observe that {\it larger versions of Whisper underperformed compared to the medium version}. This was particularly evident on the age category, where {\it all models behaved consistently better for higher age groups}.
As for the carbon footprint at inference, we observed clear disparities between the different carbon tracking tools. Although all show similar measurements for both datasets, {\it eco2ai consistently underestimated carbon emissions}, compared to both carbontracker and code carbon. 
Also, when refining by GPU and configuration,  {\it wide GPU bandwith seems to have a positive impact in both carbon emissions and energy comsumption.}




\section{Dataset Description}\label{Sec.2}


Two datasets were selected to investigate biases in terms of age, gender, and accent: the Artie-Bias on read speech, and the CCv2  targeting spontaneous speech.


\subsection{Artie-Bias Dataset}


The Artie Bias dataset \cite{artiebias} is based on the test set partition of the English Common Voice corpus, released in June 2019, totaling 1712 utterances (approx 2.4 hours), in reading mode. 
Demographic information for each speaker includes gender (3 categories), age (8 groups), and accent (17 English accents).  We have detailed the speech utterance count distribution across gender, accent, and age in Fig 1.a. The corpus is not balanced in terms of gender, as male speakers account for 1,431 utterances, whereas female speakers total only 257. Due to the skewed representation of accents, we only consider accents from the United States (US), India, and United Kingdom (UK). Similarly, we only take into account 6 age group intervals varying from less than twenty ($<20$) to sixty and above ($\geq 60$) in 10-year groups.

\subsection{Casual conversation dataset version 2}

In 2023, Meta AI published the casual conversation dataset version 2 (CCv2) \cite{ccdv2,ccdv2other}, a fairness-centric self-recorded multilingual videos from different demographic world regions, with 5567 unique speakers. The dataset includes various self-labeled attributes such as gender, age, accent, location, skin tone, voice timbre, etc. In this study, we focused investigation only on gender, age, and accent for the English language. We have illustrated the sample distribution for various attributes in Fig 1.b. on a total of 1053 speech utterances. In the CCv2 dataset, the textual content of all the speech utterances remains the same across speakers from all the demographics. This allows us to observe the impact of biases without considering the pertaining variations in performances due to textual content. For accent bias, we only considered accent variants from US English speakers, Indonesia, Brazil, and Mexico. Furthermore, we used the same age-group intervals as used in the Artie Bias dataset.

\begin{figure*}[!t] 
  \centering
  \includegraphics[scale=0.17]{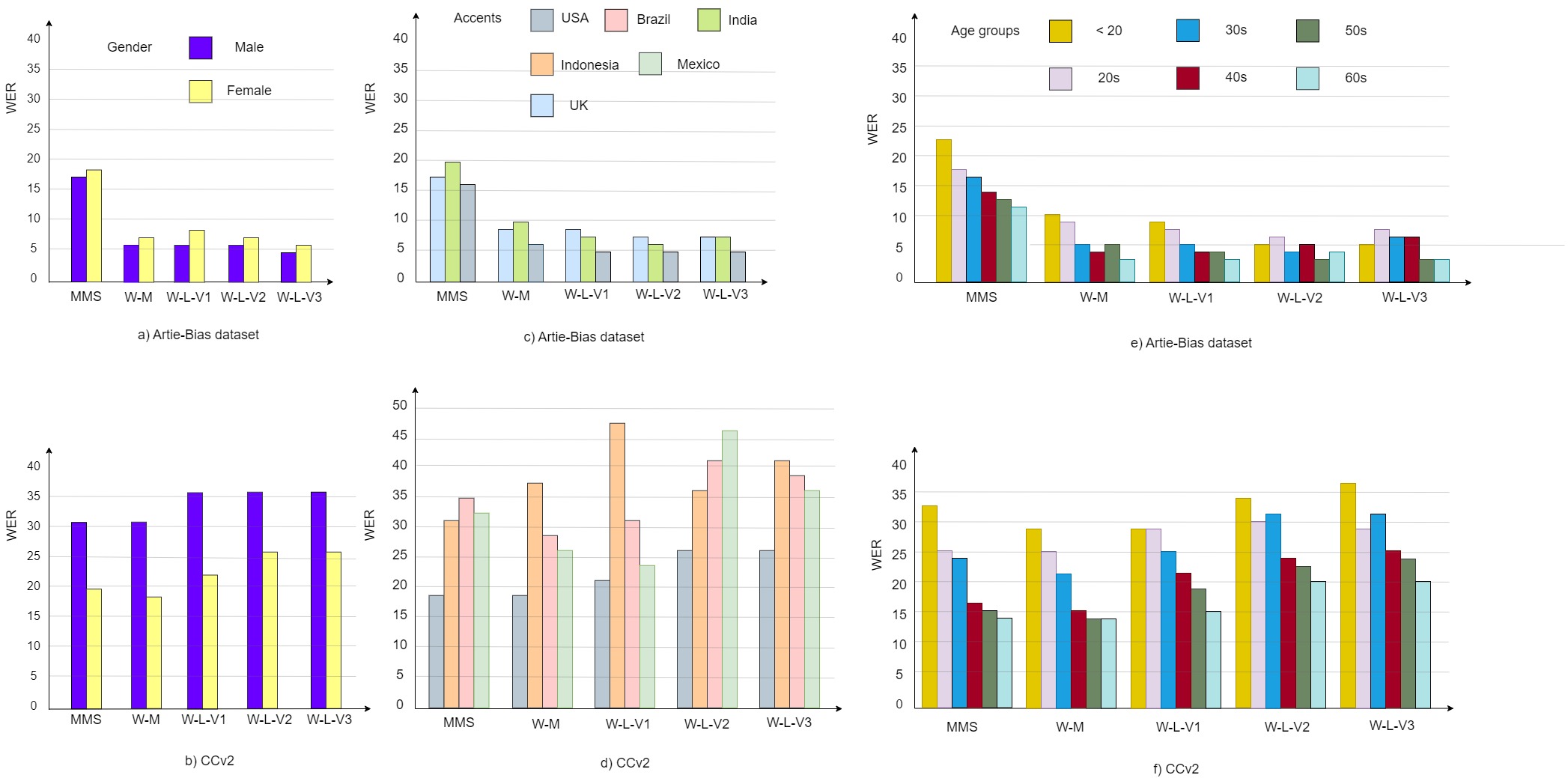} 
  \caption{Bar plots depicting Whisper and MMS ASR performances across gender, accent,  and age. Whisper ASR variants are indicated respectively as  Whisper-Medium (W-M), Whisper-Large (W-L), Whisper-Large-V2 (W-L-V2), and Whisper-Large-V3 (W-L-V3).}
  \label{fig:tsne}
\end{figure*}

\section{ASR Systems}\label{Sec.3}

To evaluate the widespread usage of ASRs in both societal and industrial applications, we focused on the MMS and Whisper ASR systems, which have become reference SOTA models.

\subsection{Massively Multilingual Speech system}

In 2023, Meta AI launched the Massively Multilingual Speech (MMS) project, as detailed in \cite{mms}, which significantly broadened its language coverage to include more than  1100 languages across various speech processing applications. The MMS project encompasses a range of tasks, including speech recognition, language identification, and speech synthesis. Built upon the wav2vec 2.0 architecture \cite{w2vec}, MMS has been trained using a combination of cross-lingual self-supervised learning and supervised pre-training for ASR. It integrates language adapters that can be dynamically loaded and swapped during inference, featuring multiple Transformer blocks, each enhanced with a language-specific adapter. The MMS system is offered in two variants based on model parameters, comprising 317 million and 965 million parameters. For this investigation, we employed the MMS system with 965 million model parameters.

\subsection{Whisper}

Whisper \cite{whisper}, introduced by OpenAI\footnote{https://openai.com/research/whisper} in 2022, is a robust speech recognition model. It is trained using multitask learning on a dataset of 680k hours of labeled multilingual recordings, 
supporting  96 languages. Whisper primarily utilizes the Transformer encoder-decoder architecture, to provide a multi-task learning framework across applications such as ASR, language identification, speech translation, etc. Whisper models are primarily categorized into two groups based on languages and tasks: English-only and multilingual models. In this study, we explore English-only variants of Whisper, including Medium (769 Million), Large-v1 (1550 Million), Large-v2 (1550 Million), and Large-v3 (1550 Million). Noticeably, the variants of Large mainly differ in the amount of training data used.


\section{Experimental Setup}\label{Sec.4}

This section presents the experimental setup to investigate bias and sustainability issues pertaining to the ASRs  of Section \ref{Sec.3}.


\subsection{Bias study} For the evaluation of both models, we use Word Error Rate (WER) for comparison purposes with the literature, and we also report on the p-values\footnote{Due to the page limit, we only provide p-values for gender groups, and provide the p-values for accent and age groups in the supplementary material along with Character Error Rate (CER) and Phoneme Error Rate (PER).} for statistical significance. These were obtained by pairwise ANOVA test with a threshold of 0.05, following the same protocol from \cite{artiebias}. This allows us to compare results objectively and to identify performance biases in the 5 ASR systems. We utilized the jiwer\footnote{\url{https://pypi.org/project/jiwer/}} library to compute WER, CER, and PER metrics, with preprocessing involving the English text normalizer from Whisper. 


\begin{figure*}[!t]
  \centering
 \includegraphics[scale=0.19]{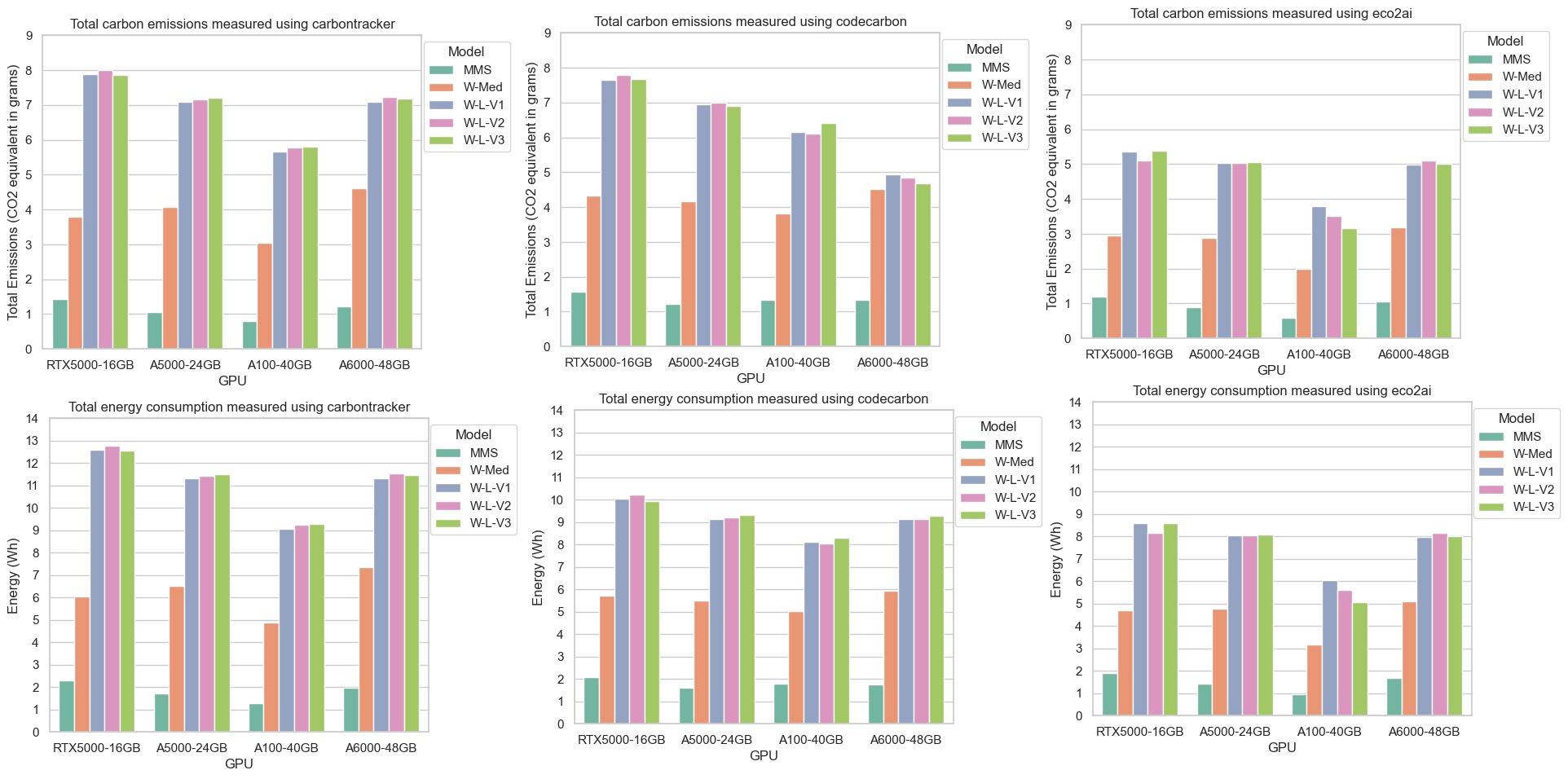}
    \caption{Bar plots depicting carbon emissions (first row) and energy consumption (second row) of MMS and  Whisper variants (W-M, W-L, W-L-V2, W-L-V3) obtained by carbontracker, codecarbon and 
 eco2ai, as described in Subsection~\ref{sustainability}. The bar clusters correspond in each bar plot correspond to the 4  NVIDIA GPUs, namely, RTX-5000-16GB, RTX-A5000-24GB, A100-48GB, and A6000-48GB. }
    \label{fig:carbon}
    \vspace{-1em}
\end{figure*}

\begin{table}[!t]
\centering
    \caption{The table provides p-values for the gender category w.r.t. all models, and both Artie-bias and CCv2 datasets, where a p-value of 0.05 or lower is statistically significant.}
    \vspace{-0.5em}
\resizebox{\columnwidth}{!}{
\begin{tabular}{|l|c|c|c|c|c|} 
\hline \textbf{Datasets} & \textbf{W-M} & \textbf{W-L-V1} & \textbf{W-L-V2} & \textbf{W-L-V3} & \textbf{MMS} \\
 \hline 

Artie-Bias		    &   0.479947	&	0.188101	&	0.238292	&	0.473756  &	0.241962 \\
CCv2	        	&   1.60E-08	&	0.00148	&	0.000121	&	0.000414  &	4.26E-06\\
\hline
\end{tabular}
}
\vspace{-1.8em}
\label{tab:stat1}
\end{table}

\subsection{Sustainability study}\label{sustainability}
We opted for 3 different platforms to measure the carbon emission intensity and energy consumption, namely, codecarbon, carbontracker, eco2ai \cite{eco2ai}. We incorporated these tools during inference of ASR on 20 mins of speech utterances across 4 NVIDIA GPUs, namely, RTX-5000-16GB, RTX-A5000-24GB, A100-40GB, and A6000-48GB. All experiments were carried out utilizing a cloud service provider based in Tamil Nadu, India, employing 32GB of RAM and 7 CPU cores. We repeated the inference run 3 times to take into account variations and utilized average estimates of energy consumption and carbon emission. System energy consumption can be quantified in either Joules (J) or watt-hours (Wh), with the latter being a unit of energy equivalent to the sustained operation of one kilowatt of power for one hour. Our study concentrated on assessing the energy usage of the GPU, CPU, and RAM, given their direct influence on the ASR inference process. Emissions across countries differ due to climate, geography, economy, fuel usage, and technological advancements. To mitigate regional differences, tools utilize emission intensity coefficients, indicating CO2 emissions per megawatt-hour of electricity. Governments incorporate these coefficients into environmental policy assessments to address emissions disparities \cite{gov_em}.

\section{Empirical Study}\label{Sec.5}



In this section, we present a comprehensive analysis on dual challenges in ASR concerning biases and sustainability. We provide detailed analyses of disparate performances across categories including gender, accent, and age groups on both read and spontaneous speech. Then, we present the empirical findings on sustainability and their analysis. 


\subsection{Results: ASR Bias}

We illustrate the ASR performances on gender category in Fig 2.a. and 2.b. for read and spontaneous speech respectively. From the bar plots, we clearly observe larger disparities in spontaneous speech across all systems, with lower performances for the male gender. Interestingly, 
this tendency is reversed on read speech, where the Whisper variants outperform the MMS. For spontaneous speech, female speech performed better than male and vice-versa for read speech, which is probably a reflection of the unbalanced training datasets across genders due to sociolinguistic differences \cite{rac1, fr_bias1}. We observed continual improvement in Whisper large-v1 to v3 variants across both datasets, but Whisper medium performed  better than other variants. 

From Table~\ref{tab:stat1}, it is also noteworthy that gender differences are not significant for read speech, unlike spontaneous speech. From Fig 2.c and 2.d, we observe subtle differences in Whisper  performances across accents in both 
datasets due to variations in model parameters. Overall, the US accent performed better than the others across both read and spontaneous speech. Noticeably, the MMS system shows higher WER in the case of read speech, while it performed better in spontaneous speech scenarios in the accent category. This illustrates the expected disparities between native and non-native English speakers. We illustrate performance disparities in ASR across the 6 age groups in Fig 2.e and 2.f for read and spontaneous speech, respectively. In the case of CCv2, the Whisper medium performed better than the other ASRs, irrespective of their model parameters. It is noteworthy that younger age groups, such as those under 20 and in their 20s, exhibit higher WERs than older age groups. 



\subsection{Results: Sustainability}
We report our findings on both carbon emissions and energy consumption obtained using 3 commonly used platforms to assess the carbon footprint of deep learning models, namely, carbontracker, codecarbon and eco2ai, in the bar plots of Fig ~\ref{fig:carbon}. We can observe a clear advantage of using MMS over the  Whisper variants throughout all platforms. As expected, we see a similar behavior of the 3 Whisper Large variants, which is due to the fact that these only differ in the amount of training data used, and thus not impacting the inference step. Also, Whisper Medium shows better performances than the Whisper Large variants, clearly due the fact that the latter have twice the number of parameters of the former. 
As for the results produced by the 3 carbon tracking platforms (Fig 4), we can observe a slightly optimistic view provided by eco2ai. However, all platforms indicate similar trends for the 5 ASR systems considered. Perhaps more interesting is the slight advantage of NVIDIA GPU A100-40GB over the other NVIDIA GPUs. This could be explained by the fact that NVIDIA GPU A100-40GB has a much  wider GPU bandwidth (1555 GB/s) than the other 3.



    


\section{Discussion and Conclusion}

We presented a thorough investigation of recently proposed  ASR systems with state-of-the-art performances, namely, the MMS and Whisper variants, in both read (Arti Bias) and spontaneous (CCv2) settings, and from bias and sustainability perspectives. 
We also compared different NVIDIA GPU architectures for their carbon footprint and reported results obtained by 3 widely used carbon tracking platforms.
Overall, as expected, ASR systems behave consistently better for read speech than for spontaneous speech, in terms of WER. 
Our findings indicate better bias performances (accent, age groups, and gender) for Whisper variants than for MMS, on read speech. However, this disparate behavior tends to disappear in spontaneous settings, where positive bias towards female speech is observed  
across all 5 ASRs considered. We also observed better ASR performances in older groups. Interestingly, larger variants of Whisper tend to behave worse than the medium counterpart.

Motivated by the widespread use of ASR systems, we also investigated carbon emissions and energy consumption in inference. Our results confirmed the superiority of MMS over the Whisper variants in terms of sustainability. 
It is noteworthy that MMS uses language-specific adapters, which restrict vocabulary output tokens, unlike the English-only variants of Whisper which have over 50K tokens. This distinction could potentially affect emission and energy consumption metrics. Indeed, language-specific adapters can help us save carbon emissions and mitigate biases. One should conduct a comprehensive analysis and perform a suitable study on each type of ASR usage.
By the comparative study of the 4 NVIDIA GPUs, we observed the impact of bandwidth on carbon footprint, with  NVIDIA GPU A100 showing better performances than the 3 other NVIDIA GPUs. It was also interesting to remark that eco2ai consistently reported lower carbon emissions and energy consumption than carbontracker and codecarbon. 
These findings emphasize the need for a comprehensive analysis of ASR systems that consider the diversity of performance metrics, implementations, and evaluation methodologies, {\it e.g.}, emission tracking and energy consumption. 

\vfill

\begin{section}{Acknowledgements}
The third named author was partially supported by the ANR
project ``Intrinsic and extrinsic evaluation of biases in LLMs'' (InExtenso), ANR-23-IAS1-0004-01. The last named author  was partially supported by the Portuguese Recovery and Resilience Plan through project C645008882-00000055
 and  by Fundação para a Ciência e Tecnologia through the INESC-ID multi-annual funding from the PIDDAC programme
(UIDB/50021/2020).

\end{section}





\bibliographystyle{IEEEtran}
\bibliography{main}

\end{document}